\documentclass[final,authoryear]{elsarticle}
\usepackage{enumitem}
\usepackage{lineno}
\usepackage{todonotes}
\modulolinenumbers[1]

\usepackage[utf8]{inputenc}
\usepackage[english]{babel}

\journal{Expert Systems with Applications}

\usepackage{tikz,bm,color}
\usetikzlibrary{fit}

\usepackage{amsmath}
\usepackage{amsthm}
\usepackage{amssymb}
\usepackage{mathtools}
\usepackage{xfrac}

\usepackage[linesnumbered, ruled, vlined]{algorithm2e}
\usepackage {listings}

\usepackage{caption}
\usepackage{subcaption}
\usepackage {float}

\usepackage{url}

\usepackage[all]{myglossary}

\usepackage{tabularx}
\usepackage{multirow}
\usepackage{booktabs}
\usepackage{array}
\usepackage{rotating}
\usepackage{soul}

\newcolumntype{L}[1]{>{\raggedright\let\newline\\\arraybackslash\hspace{0pt}}m{#1}}
\newcolumntype{C}[1]{>{\centering\let\newline\\\arraybackslash\hspace{0pt}}m{#1}}
\newcolumntype{R}[1]{>{\raggedleft\let\newline\\\arraybackslash\hspace{0pt}}m{#1}}

\theoremstyle{definition} 

\begin{document}

\nolinenumbers

\begin{frontmatter}


\title{A revision on Multi-Criteria Decision Making methods for Multi-UAV Mission Planning Support}


\author[iks]{Cristian Ramirez-Atencia}
\ead{cristian.ramirez@ovgu.de}

\author[upm]{Victor Rodriguez-Fernandez}
\ead{victor.rodriguezf@uam.es}

\author[upm]{David Camacho}
\ead{david.camacho@upm.es}

\address[iks]{Faculty of Computer Science, Otto von Guericke University Magdeburg, Germany}
\address[upm]{Departamento de Sistemas Inform\'aticos, Universidad Polit\'ecnica de Madrid, Spain}




\begin{abstract}
Over the last decade, \glspl{uav} have been extensively used in many commercial applications due to their manageability and risk avoidance. One of the main problems considered is the Mission Planning for multiple \glspl{uav}, where a solution plan must be found satisfying the different constraints of the problem. Moreover, this problem has some variables that must be optimized simultaneously, such as the makespan, the cost of the mission or the risk. Due to this, the problem has a lot of possible optimal solutions, and the operator must select the final solution to be executed among them. In order to reduce the workload of the operator in this decision process, a \gls{dss} becomes necessary. In this work, a \gls{dss} consisting of ranking and filtering functions, which order and reduce the optimal solutions, has been designed. 
A wide range of \gls{mcdm} methods, including some fuzzy \gls{mcdm}, are compared on a multi-UAV mission planning scenario, in order to study which method could fit better in a multi-UAV decision support system. Expert operators have evaluated the solutions returned in order to score and compare fuzzy-based methods against classical ones, and the results show that fuzzy methods generally achieve better average scores. On the other hand, for the filtering system, a similarity function based on the proximity of the solutions has been designed, and the threshold used for the filtering is studied.
\end{abstract}


\end{frontmatter}

\glsresetall

\section{Introduction}\label{introduction}
The fast development of \glspl{uav} capabilities in the last years has led to an upswing in new military and commercial applications, where many fields such as surveillance~\citep{Gu2018Multiple}, training~\citep{Rodriguez-Fernandez2018Automatic}, fire fighting~\citep{Ghamry2017Multiple}, agriculture~\citep{Tokekar2016Sensor} or disaster and crisis management~\citep{Nedjati2016Post} have been researched.

One of the main research fields of \glspl{uav} is mission planing, where a collaborative swarm of \glspl{uav} must perform some tasks taking into account some spatial and time constraints. It is a highlight goal in \gls{uav} research, since it is a complex problem that hardens \gls{uav} operators workload. Nowadays, \glspl{uav} are controlled remotely by human operators from \glspl{gcs}, using rudimentary planning systems, such as pre-configured plans, manually provided schedules or classical planners. These classical planners, based on a graph search or a logic engine, usually undergo severe practical limitations and their solvers have a high computational cost. 

In order to undertake complex coordinated missions, planning systems demand more efficient problem-solving capabilities to cope with conflicting objectives and stringent constraints over both spatial and time domains. In addition, these problems usually aim to optimize several conflicting objectives, including the fuel consumption, the makespan, the cost of the mission, the number of \glspl{uav} to employ and different risk factors that could compromise the mission, among others. Therefore, the so-called \gls{mcmpp} have lately gained momentum in the research community, propelled by a notable increase in the scales and number of applications foreseen for \glspl{uav} in the near future~\citep{Ramirez-Atencia2017New}.

Nowadays, in most \glspl{gcs}, there are several operators that usually deal with just one \gls{uav}, due to the high complexity of a mission. One of the main aims in this field is to reduce this complexity, so this relation can be swapped, and in a near future a single operator would control multiple vehicles. For this, one of the principal requirements is to automatize and reduce the complexity of the \gls{mcmpp}. In previous works, the \gls{mcmpp} has been formulated as a \gls{csp}~\citep{Ramirez-Atencia2018Constrained}. In that work, the variables of the problem are the assignments of \glspl{uav} to tasks, \glspl{gcs}, sensors and further variables, and several spatial and temporal constraints are defined to assure the consistency of the solutions. Then, the problem has been solved using a \gls{moea}~\citep{Ramirez-Atencia2018Weighted}, considering seven conflicting objectives. In this algorithm, an estimation of the \gls{pof} is inferred so as to get a portfolio of solutions (mission plans) differently albeit optimally balancing the considered conflicting objectives.

One critical point in this problem, and the motivation of this work, is that sometimes the entire \gls{pof} comprises a large number of solutions. In this situation, the process of decision making to select one solution among them becomes a difficult task for the operator. In some cases, the operator can provide a priori information about his/her preferences, which can be used in the optimization process. However, most times the operator does not provide this information, so a posteriori approaches must be considered. In this case, the a posteriori multi-objective optimization algorithm will most times return a large number of solutions, and the decision process becomes highly complex. 

Such was the case in~\citep{Ramirez-Atencia2018Weighted}, where a large number of non-dominated solutions were obtained. In order to reduce the number of solutions, in~\citep{Ramirez-Atencia2017Knee}, a Knee-Point \gls{moea} approach was used to only return the most significant ones according to the optimization variables, i.e. the knee points of the \gls{pof}. Although this approach highly reduces the number of solutions (from hundreds to tens), in most cases there will still be several solutions and the operator has to take the final decision of which one is the most appropriate for the mission. This decision process is not easy, as each solution comprises several variable assignments and several criteria defining its quality.

In this context, a \gls{dss} becomes essential for guiding the operator in the process of selecting the best solution. This \gls{dss} provides a ranking system that will order the solutions obtained by the optimization algorithm, and, additionally, a filtering system that reduces the number of solutions provided. There exists several \gls{mcdm} methods that can be used to rank and select the solutions based on the operator preferences, which are used as percentages of importance or orders of the different criteria. Nevertheless, sometimes these preferences may be fuzzy, so they must be considered in a different way. For these cases, there exist fuzzy versions of the \gls{mcdm} methods, where these preferences are treated as fuzzy numbers.

The goal of this paper is to develop a \gls{dss} to assist \gls{uav} operators in the selection of the mission plan. In a previous work~\citep{Ramirez-Atencia2018Multi}, some classical \gls{mcdm} methods where used to rank the solutions of the \gls{mcmpp} returned by a \gls{moea}~\citep{Ramirez-Atencia2017Knee}. In this paper, new \gls{mcdm} methods are considered, as well as their fuzzy versions. As the preferences provided by operators are also fuzzy, it is expected for these methods to provide better results. A number of realistic mission scenarios is used to test these methods. Expert operators evaluated the solutions provided for each mission according to some operator profiles (preferences), and through this evaluation, the \gls{mcdm} algorithms have been scored. In addition, a filtering system that omits similar solutions has been designed and tested in order to provide a complete Multi-Criteria \gls{dss}.

The rest of the paper is structured as follows. Section \ref{dss} presents some background on \gls{dss}, including Fuzzy \gls{mcdm}. Section \ref{dssmpp} describes the \gls{uav} mission planning problem and the \gls{dss} developed to help the operator in the selection of the best solutions, including the ranking and filtering systems. Section \ref{experiments} presents the results obtained in the experiments for both the ranking and filtering systems. Finally, the last section draws conclusions and outlines future research lines related to this work.

\section{Background on Decision Support Systems}\label{dss}
\Glspl{dss}~\citep{Shim2002,Fulop2005} are information systems used in decision making and problem solving. Research on \gls{dss} is focused on the efficiency of user decision making and how to increase the effectiveness of that decision.  
One of the most common fields in \gls{dss} is \gls{mcdm}~\citep{Triantaphyllou2000}, where multiple criteria $f_1, f_2, .. f_n \in F$ are taken into account for each of the alternatives $s_1, s_2, ... s_m \in S$ considered. The problems of \gls{mcdm} appear and are intensely applied in many domains, such as Economics, Social Sciences, Medical Sciences etc. 

There are two main subfields in \gls{mcdm}: \gls{modm}, focused on design problems, and \gls{madm}, focused on evaluation problems. In \gls{modm} alternatives are not explicitly known, but instead a mathematical model is used to find them. In most cases, there is an infinite or very large number of alternatives, and the methods in this subfield are mostly the same as in Multi-Objective Optimization.

On the other hand, in \gls{madm}~\citep{Triantaphyllou2001}, there is a finite number of alternatives, represented by their performance in multiple (conflicting) criteria. The decision makers have to provide some weights or preferences for each of these criteria (e.g. percentage numbers, degrees of importance, etc.). The main goals in these problems are choice, sorting or ranking~\citep{Zavadskas2006Evaluation}. In choice problems, one or a few incomparable alternatives are selected as the best alternative. In sorting problems, alternatives are grouped into predefined categories, with similar behaviors or characteristics. In ranking problems, alternatives are ordered from best to worst according to some scores or pairwise comparisons. 
For this last kind of problems, there exist several methods, such as the following:

\begin{itemize}
\item \textbf{\Gls{wsm}}~\citep{Triantaphyllou2000}. This approach consists of an utility function with weight values $w_f$ (in the interval $[0,1]$) for each criteria or factor $f \in F$, according to their importance. Given the values $v_f(s)$ for each factor in an alternative or solution $s \in S$, the utility function for \gls{wsm} can be expressed as a weighted sum:

\begin{equation}
rank_{value} (s) = \sum_{f \in F} \frac{w_f \times v_f(s)}{\max\limits_{s' \in S} v_f(s')}
\end{equation}

All of these $rank_{value}(s)$ are then used to rank the alternatives.

\item \textbf{\Gls{wpm}}~\citep{Tofallis2014}. This method is similar to \gls{wsm}, but instead of addition, each decision alternative is compared with the others by multiplying a number of ratios, one for each decision criterion. Each ratio is raised to the power equivalent to the relative weight of the corresponding criterion. In order to compare two solutions $s_i$ and $s_j$, whose criteria values are $v_f(s_i)$ and $v_f(s_j)$, then the following product is considered:

\begin{equation}
P(\sfrac{s_i}{s_j}) = \prod_{f \in F} \left(\frac{v_f(s_i)}{v_f(s_j)}\right)^{w_f}
\end{equation}

where the ratio $P(\sfrac{s_i}{s_j}) \geq 1$ indicates that alternative $s_i$ is better.

\item \textbf{\gls{ahp}}~\citep{Saaty2001}. This method allows individuals or groups to make complex decisions. The procedure of \gls{ahp} firstly decomposes the decision problem into a hierarchy of more easily comprehended sub-problems, and then analyse each of them independently, comparing each pair of elements of the hierarchy. In most problems, there would be three levels in this hierarchy, in the following top-down order: the goal, the criteria and the alternatives. The core concept of \gls{ahp} is that alternatives are always compared pairwise. For this, the decision makers must provide the weights as pairwise comparisons of the criteria (e.g. using a scale 1-9 where 1 means equally important and 9 means the first is extremely important compared to the second), resulting in a matrix $A$ of weight comparisons. Also, for each criteria, alternatives are similarly compared pairwise according to that criterion.

All of these comparisons are converted to numerical values, providing a numerical weight or priority for each element of the hierarchy. Several methods can be used in this process (e.g. approximation method, geometric mean, etc.), being the Eigenvalue method the most common. In this method, the priorities $p$ of each alternative/criteria are computed as the eigenvectors of the following equation:

\begin{equation}
    A \cdot p = \lambda \cdot p
\end{equation}

where $\lambda$ is the eigenvalue, i.e. the maximum value satisfying the equation. Then, based on these priorities on each level of the hierarchy, numerical priorities can be calculated for each of the decision alternatives through additive aggregation (or in some cases multiplicative aggregation):

\begin{equation}
    P_s = \sum_{f \in F} w_f \cdot p_{sf}
\end{equation}

where $P_s$ is the global priority of alternative $s$, $w_f$ is the priority of criterion $f$ and $p_{sf}$ is the priority of alternative $s$ w.r.t criterion $f$. Based on these global priorities, alternatives are ranked.

\item \textbf{VIKOR}~\citep{Opricovic2004}. This method solves decision problems with conflicting and noncommensurable (different units) criteria. Assuming that compromise is acceptable for conflict resolution, the decision maker looks for a compromise solution $F^C$ that is the closest to the ideal $F^*$, and the alternatives are evaluated according to all established criteria. In its process, VIKOR determines the best $f^*$ and the worst $f^-$ values for every criteria $f \in F$. Then, the weighted and normalized Manhattan ($S_s$) and Chebyshev ($R_s$) distances are computed for every alternative:

\begin{align}
S_s = \sum_{f \in F} \frac{w_f \cdot (f^* - v_f(s))}{f^* - f^-}\label{eq:manhattanvikor}\\
R_s = \max_{f \in F} \frac{w_f \cdot (f^* - v_f(s))}{f^* - f^-}\label{eq:chebyshevvikor}
\end{align}

From these distances, the best $S^*=\min_k S_k$, $R^*=\min_k R_k$ and worst $S^-=\max_k S_k$, $R^-=\max_k R_k$ values are used to compute the maximum group utility (or majority of the criteria):

\begin{equation}
Q_s = v \cdot \frac{S_s - S^*}{S^- - S^*} + (1-v) \cdot \frac{R_s - R^*}{R^- - R^*}
\end{equation}

where $v$ is the weight of the strategy of the maximum group utility, which could be compromised by $v=0.5$. Following the method, VIKOR rank the alternatives by sorting the values of $S$, $R$ and $Q$ and proposes a compromise solution as the best value of $Q$ given some conditions. For ranking problems, $Q$ ranking can be used.



\item \textbf{\Gls{topsis}}~\citep{Garcia-Cascales2012}. It is a method of compensatory aggregation that compares a set of alternatives by identifying weights for each criterion, normalising scores for each criterion and calculating the geometric distance between each alternative and the ideal alternative, which is the best score in each criterion. An assumption of \gls{topsis} is that the criteria are monotonically increasing or decreasing. The normalisation can be performed using different procedures, such as vectorial (distributive) normalisation:

\begin{equation}\label{eq:vectornorm}
    r_{sf} = \frac{v_f(s)}{\sqrt{\sum\limits_{s' \in S} v_f(s')^2}}
\end{equation}

or linear transformation of maximum (also called ideal normalization):

\begin{equation}\label{eq:linearnorm}
    r_{sf} = \left\{ 
\begin{array}{ll}
  \frac{v_f(s)}{\max\limits_{s' \in S} v_f(s')} & \text{if criterion } f \text{ is maximized}\\
  \frac{\min\limits_{s' \in S} v_f(s')}{v_f(s)} & \text{if criterion } f \text{ is minimized}
 \end{array}\right.
\end{equation}

These normalized scores are then multiplied by the corresponding weights $w_f$, and compared against the ideal and anti-ideal solutions. Virtual ideal and anti-ideal solutions are constructed as the best and worst scores on each criterion, respectively. Then, distance to the ideal $d_s^{+}$ and anti-ideal $d_s^{-}$ solutions are computed for each solution $s$ using the Euclidean distance. Finally, the closeness ratio for each alternative is computed as:

\begin{equation}\label{eq:closenessratio}
    C_s = \frac{d_s^{-}}{d_s^{+}+d_s^{-}}
\end{equation}

This closeness ratio establishes the ranking, where the highest value indicates the preferred alternative.

\item \textbf{ELECTRE}~\citep{Roy1991}. It is an outranking method that compares all feasible alternatives by pair building up some binary relations, crisp or fuzzy, and then exploit in appropriate way these relations in order to obtain final recommendations. It is commonly used when there are more than two criteria to avoid the compensation effect of weights. Several ELECTRE methods were proposed for different kinds of problems, being ELECTRE II, ELECTRE III and ELECTRE IV proposed for ranking problems. ELECTRE III consists of two phases: first it constructs one or several outranking relations, which aims at comparing in a comprehensive way the alternatives, and then an exploitation procedure elaborates preference order using these outranking relations. Criteria in ELECTRE III have four distinct sets of parameters: the criteria weights ($w$), and the indifference ($q$), the preference ($p$) and the veto ($v$) thresholds. These thresholds indicate the largest difference between the performance of alternatives w.r.t a given criterion such that they remain indifferent ($q$), one is preferred over the other ($p$) or the `discordant' difference in favour of one option negates any possible outranking relationship indicated by other criteria ($v$).

The outranking relation `$a$ outranks $b$', meaning that $a$ is at least as good as $b$, is defined through the outranking degree $S(a,b) \in [0,1]$, where a value closer to 1 indicates a stronger assertion. This outranking degree $S(a,b)$ is computed by aggregating two perspectives: concordance degree (using the indifference and preference thresholds) and discordance degree (using the veto threshold). 
For every criterion, partial concordance degree $c_f(a,b)$ is computed using indifference $q_f$ and preference $p_f$ thresholds through linear interpolation, where if alternatives $b$ is indifferent or worse than $a$ according to $q_f$, then $c_f(a,b)=1$, while if $b$ is preferred to $a$ according to $p_f$, then $c_f(a,b)=0$. Then, the global concordance degree $C(a,b)$ is computed as a weighted sum of partial concordance degrees. 
Partial discordance degree $d_f(a,b)$ is computed using veto threshold $v_f$ through linear interpolation, where if alternatives $b$ is better than $a$ according to $v_f$ (i.e. strong disagreement with the assertion), then $d_f(a,b)=1$, while if $a$ is as least as good as $b$ according to $p_f$ (i.e. no reason to refute the assertion), then $d_f(a,b)=0$. 
Then, the outranking degree is computed using the concordance and discordance degrees as:

\begin{equation}
    S(a,b) = C(a,b) \cdot \prod_{f \in V} \left[ \frac{1-d_f(a,b)}{1-C(a,b)} \right]
\end{equation}

where $V$ is the set of criteria for which $d_f(a,b)>C(a,b)$. All the values $S(a,b)$ for each pair of alternatives provides the credibility matrix, from where the preference relation `$a$ is preferred over $b$', $a P b$, is expressed by combining $S(a,b)$ and $S(b,a)$ given a cut-off level $\lambda$.

The exploitation procedure consists of ascending and descending distillation procedures that lead to two transitive pre-order, whose intersection generates the final ranking. These distillations are based on qualification scores of alternatives: 

\begin{equation}
    Score(a) = \#\{b|a P b\}-\#\{b|b P a\}
\end{equation}

The descending distillation procedure extract the best alternative(s) according to these scores, then it repeats the same process with the remaining alternatives but decreasing the cutoff level $\lambda$ (so it becomes easier for alternatives to be preferred) until remaining alternatives have the same scores. The ascending distillation procedure is similar, but extracting the worst alternative(s). The final partial pre-order $O$ is defined as the intersection of descending pre-order $O_1$ and ascending pre-order $O_2$.

\item \textbf{\gls{mmoora}}~\citep{Brauers2010}. This method starts with a decision matrix showing the performance of different alternatives with respect to various attributes (objectives), and uses three ranking systems that are then combined. First, a ratio system is developed in which each performance of an alternative w.r.t each criterion is normalized as in Equation \ref{eq:vectornorm} and then all the criteria for each alternative are aggregated as:

\begin{equation}
    y_s^* = \sum_{f \in F^{max}} w_f \cdot r_{sf} - \sum_{f \in F^{min}} w_f \cdot r_{sf}
\end{equation}

where $F^{max}$ and $F^{min}$ are the criteria to be maximized and minimized, respectively. With this, the alternatives are ranked in descending order of $y_s^*$ values, resulting in the ratio system ranking.

Secondly, a reference point is computed as the best values of the normalized performance of alternatives w.r.t. each criterion:

\begin{equation}
    r_{f} = \left\{ 
\begin{array}{ll}
  \max\limits_{s' \in S} r_{s'f} & \text{if } f \in F^{max}\\ 
  \min\limits_{s' \in S} r_{s'f} & \text{if } f \in F^{min}\\ 
 \end{array}\right.
\end{equation}

The distance between each normalized alternative and this reference point are measured using Tchebycheff Min-Max metric:

\begin{equation}
    \min\limits_{s \in S} \left\{ \max\limits_{f \in F} |w_f \cdot r_{f} - w_f \cdot r_{sf}| \right\}
\end{equation}

Through this metric, the reference point system ranking is obtained. 

The third ranking system consists of a full multiplicative form, where the utility of an alternative $s$ is given by:

\begin{equation}\label{eq:mmooramult}
U_s = \frac{A_s}{B_s} = \frac{\prod\limits_{f \in F^{max}} v_f(s)}{\prod\limits_{f \in F^{min}} v_f(s)}
\end{equation}

Finally, these three ranking are combined based on dominance theory~\citep{Brauers2014} to get the final ranking for \gls{mmoora}.

\item \textbf{\Gls{rim}}~\citep{Cables2016}. This method evolves a value or set of values (the reference ideal), that is always maintained between a maximum value and a minimum value. This method is characterized to be independent of the type of data, and does not present rank reversal, an aspect that is not present in other \gls{mcdm}. \Gls{rim} requires for each criterion the range $[A,B]$ that belongs to a universe of discourse and the reference ideal $[C,D]$ that represents the maximum importance or relevance in a given range. The normalization based on these parameters is computed as: 

\begin{equation}
    y_{sf} = \left\{ 
\begin{array}{ll}
  1 & \text{if } v_f(s) \in [C,D]\\ 
  1 - \frac{min(|v_f(s)-C|,|v_f(s)-D|)}{|A-C|} & \text{if } v_f(s) \in [A,C] \wedge A \neq C\\ 
  1 - \frac{min(|v_f(s)-C|,|v_f(s)-D|)}{|D-B|} & \text{if } v_f(s) \in [D,B] \wedge D \neq B\\  
 \end{array}\right.
\end{equation}

Then, the weighted normalized values $y_{sf}' = w_f \cdot y_{sf}$ are used to calculate the variation to the normalized reference ideal (and anti-ideal) for each alternative:

\begin{align}
I_s^+ = \sqrt{\sum\limits_{f \in F} (y_{sf}'-w_f)^2}\\
I_s^- = \sqrt{\sum\limits_{f \in F} (y_{sf}')^2}
\end{align}

Finally, the relative index $R_s$ of each alternative is computed using these values and Equation \ref{eq:closenessratio}, similarly as in \gls{topsis}. The alternatives are ranked in descending order according to their relative index.

\item \textbf{\Gls{waspas}}~\citep{Zavadskas2012Optimization}. It is an aggregation method that applies a joint criterion based on the additive relative importance (computed using \gls{wsm}) and the multiplicative relative importance (computed using \gls{wpm}), supposing the increase of ranking accuracy and, respectively, the effectiveness of decision making. The method performs a normalization similar to ideal normalization in \gls{topsis} (Equation \ref{eq:linearnorm}). The joint generalized criterion $Q_s$ is then computed as:

\begin{equation}
Q_s = \lambda \sum_{f \in F} w_f \cdot r_{sf} + (1-\lambda) \prod_{f \in F} (r_{sf})^{w_f}
\end{equation}

where $\lambda$ indicates the trade-off between the models (usually $\lambda=0.5$). These joint generalized criterion values are ranked in a descending order, and the highest amount of joint generalized criterion has the highest rank.
\end{itemize}

\subsection{Fuzzy MCDM}
Sometimes, the preferences provided by the decision makers are not smooth, as the weights of the criteria could be expressed according to their importance using some linguistic term (e.g. low, medium, high). As these linguistic terms define a fuzzy degree of importance, they could be expressed as fuzzy numbers, and new fuzzy versions of the \gls{mcdm} methods must be employed to consider them.

Designing ranking for each option is difficult for a decision maker under operational conditions, but fuzzy methods enable a decision maker to use fuzzy numbers in place of accurate ones. One of the most common fuzzy numbers are \glspl{tfn}\citep{Borovivcka2014}, which are useful to handle imprecise numerical quantities. A \gls{tfn} can be defined as a triplet $(a_1, a_2, a_3)$, where $a_2$ is the value of maximum membership to the fuzzy set (i.e. most promising value) and $a_1$ and $a_3$ are the limits of membership to the fuzzy set (i.e. smallest and largest possible values).

\begin{figure}[!h]
\begin{center}
\includegraphics[width=\textwidth]{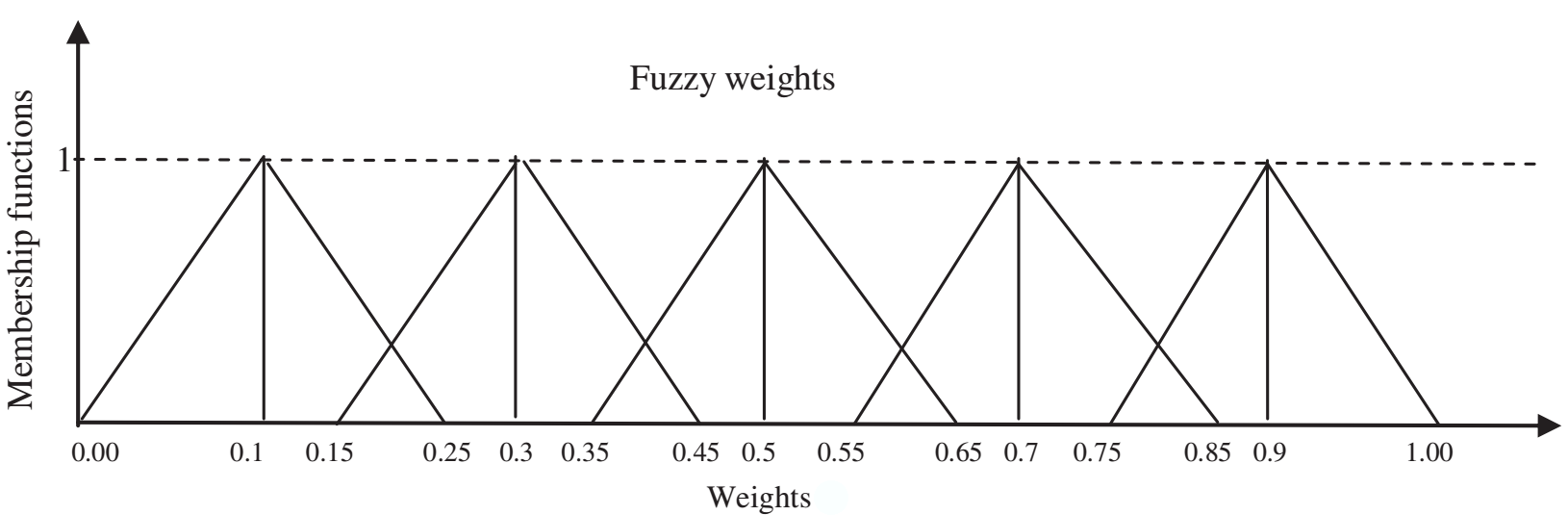}
\caption{Linguistic scales and fuzzy weights numbers adopted.}
\label{fig:fuzzy_weights}
\end{center}
\end{figure}

In order to convert the linguistic terms to \glspl{tfn}, a conversion
scale is used (see Figure \ref{fig:fuzzy_weights}), where both the performance score and membership function $\mu_A$ are in the range $[0,1]$. The value $0$ means that the element is not a member of the fuzzy set, while the value 1 means that the element is fully a member of the fuzzy set. Values between 0 and 1 characterize fuzzy members, which belong to the fuzzy set only partially.

Algebra operations on \gls{tfn} are similar to tuple operations for the sum and multiplication of \glspl{tfn}, as well as the multiplication of a constant crisp number with a \gls{tfn}. But in the case of subtraction and division, these operations for \glspl{tfn} $\tilde{a}=(a_1,a_2,a_3)$ and $\tilde{b}=(b_1,b_2,b_3)$ are defined as follows:
 
\begin{align}
    \tilde{a} \ominus \tilde{b} &= (a_1-b_3,a_2-b_2,a_3-b_1)\label{eq:fuzzydiff}\\
    \tilde{a} \oslash \tilde{b} &= (\frac{a_1}{b_3},\frac{a_2}{b_2},\frac{a_3}{b_1})\label{eq:fuzzydiv}
\end{align}

On the other hand, the distance between two \gls{tfn} can be computed using the vertex method:

\begin{equation}\label{eq:fuzzydist}
    d(\tilde{a}, \tilde{b})=\sqrt{\frac{1}{3}[(a_1-b_1)^2+(a_2-b_2)^2+(a_3-b_3)^2]}
\end{equation}
 
There exist some fuzzy versions of \gls{mcdm} methods that deal with \glspl{tfn}, including:

\begin{itemize}
    \item \textbf{Fuzzy \gls{ahp}}~\citep{Krejci2017}. In this fuzzy version, pairwise comparisons are now expressed using linguistic variables (specified by \glspl{tfn}). As in \gls{ahp}, comparisons are converted into (fuzzy) numerical weights. In this case, the fuzzy version of the geometric mean is easier to apply than the Fuzzy Eigenvalue method. Given the \gls{tfn} values $\tilde{a}_{ij}=({a_{ij}}_1,{a_{ij}}_2,{a_{ij}}_3)$ in matrix $A$ of size $p \times p$, the weights $\tilde{w}_{i}=({w_i}_1,{w_i}_2,{w_i}_3)$ can be computed as:
    
    \begin{align}
        {w_i}_1 = \frac{\sqrt[p]{\prod_{j=1}^p {a_{ij}}_1}}{\sum_{k=1}^p \sqrt[p]{\prod_{j=1}^p {a_{kj}}_3}} \\
        {w_i}_2 = \frac{\sqrt[p]{\prod_{j=1}^p {a_{ij}}_2}}{\sum_{k=1}^p \sqrt[p]{\prod_{j=1}^p {a_{kj}}_2}} \\
        {w_i}_3 = \frac{\sqrt[p]{\prod_{j=1}^p {a_{ij}}_3}}{\sum_{k=1}^p \sqrt[p]{\prod_{j=1}^p {a_{kj}}_1}}
    \end{align}

    Then, (fuzzy) numerical priorities are computed similarly to \gls{ahp} using \gls{tfn} algebra operations. Finally, to rank these \glspl{tfn}, different fuzzy number comparators can be used, such as Chen method~\citep{Chen1995}, which is used in this work.
    
    \item \textbf{Fuzzy VIKOR}~\citep{Opricovic2011}. This fuzzy extension uses the sames methods as in classic VIKOR for the computation of weighted and normalized Manhattan $\tilde{S}_s$ and Chebyshev $\tilde{R}_s$ distances based on the best $\tilde{f}^*=(f^*_1,f^*_2,f^*_3)$ and worst $\tilde{f}^-=(f^-_1,f^-_2,f^-_3)$ value for each criterion $f \in F$. For this fuzzy version, when performing operations with fuzzy numbers, Equations \ref{eq:manhattanvikor} and \ref{eq:chebyshevvikor} must take into account the normalized fuzzy difference:
    
    \begin{equation}
        \tilde{d}_f(\tilde{s}) = \left(\frac{f^*_1 - v_f(s)_3}{f^*_3 - f^-_1}, \frac{f^*_2 - v_f(s)_2}{f^*_3 - f^-_1}, \frac{f^*_3 - v_f(s)_1 }{f^*_3 - f^-_1}\right)
    \end{equation}
    
    Similarly, when computing the maximum group utility $\tilde{Q}_s$, the denominators become the difference between the largest value of the best and the smallest value of the worst; while numerators must consider consistently any fuzzy subtraction as stated in Equation \ref{eq:fuzzydiff}. Finally, \glspl{tfn} are defuzzified using the method `2nd weighted mean', where a crisp value for the \gls{tfn} is approximated as:
    
    \begin{equation}
        c(a) = \frac{a_1+2a_2+a_3}{4}
    \end{equation}
    
    and these crisp values are used as scores for ranking the same way as in VIKOR.
    
    \item \textbf{Fuzzy \gls{topsis}}~\citep{Chen20000}. The extension of \gls{topsis} for \gls{tfn} performs vectorial normalization as in Equation \ref{eq:vectornorm} taking into account the fuzzy operations. In the case of linear normalization, Equation \ref{eq:linearnorm} is extended for fuzzy numbers, and in this case the maximum operator returns the largest value of the \gls{tfn} (i.e. $a_3$) and the minimum operator returns the smallest value of the \gls{tfn} (i.e. $a_1$). Fuzzy ideal and anti-ideal points can be constructed by defining fuzzy numbers $v_f^*=(1,1,1)$ and $v_f^-(0,0,0)$ for each criterion, respectively, and the distances of solutions to these ideal $d_s^{+}$ and anti-ideal $d_s^{-}$ points are computed using the fuzzy distance provided in Equation \ref{eq:fuzzydist}. With this, a crisp value is obtained, and so closeness ratio can be applied to get the ranking of solutions.

    \item \textbf{Fuzzy \gls{mmoora}}~\citep{Balezentis2013}. In this fuzzy extension, each of the three ranking methods is converted into its fuzzy version by considering \gls{tfn} and fuzzy operations. In the ratio system, fuzzy distance from Equation \ref{eq:fuzzydist} is used in the normalization of the performance of alternatives w.r.t. each criterion, and the obtained results after applying this method are defuzzified using the best nonfuzzy performance  (BNP) value:
    
    \begin{equation}
        BNP(a)=\frac{(a_3-a_1) + (a_2-a_1)}{3} + a_1
    \end{equation}
    
    For the reference point method, fuzzy distance is used to measure the distance of each solution to the ideal point, and the resulting crisp values are then used in the Tchebycheff Min-Max metric to obtain the reference point ranking. The fully multiplicative form follows Equation \ref{eq:mmooramult} considering fuzzy operations, and then the fuzzy numbers $\tilde{U}_s$ are defuzzified using BNP. Finally, these three ranking are combined using dominance theory as in classic \gls{mmoora}.
    
    \item \textbf{Fuzzy \gls{waspas}}~\citep{Turskis2015}. The extension of \gls{waspas} for supporting fuzzy numbers is straightforward by considering fuzzy operations in the \gls{wsm} and \gls{wpm} formulas. It must be noticed here, that the exponentiation operation of \glspl{tfn} $\tilde{v}={v_1,v_2,v_3}$ and $\tilde{w}={w_1,w_2,w_3}$ for \gls{wpm} is computed as follows:
    
    \begin{equation}
        \tilde{v}^{\tilde{w}} = ({v_1}^{w_3},{v_2}^{w_2},{v_3}^{w_1})
    \end{equation}
    
    Then, the \glspl{tfn} obtained for \gls{wsm} and \gls{wpm} are defuzzified using the centroid method:
    
    \begin{equation}
        c(a) = \frac{a_1+a_2+a_3}{3}  
    \end{equation}
    
    and the rest of the method is applied with the obtained crisp numbers.

\end{itemize}

\section{A DSS for Mission Plan Selection}\label{dssmpp}
The \gls{mcmpp} considered in this work has been defined in previous works~\citep{Ramirez-Atencia2018Constrained} as a $\texttt{T}$-sized set of \textit{tasks} $\mathcal{T}\doteq\{T_1,T_2,\ldots,T_\texttt{T}\}$ to be performed by a swarm of $\texttt{U}$ \glspl{uav} $\mathcal{U}=\{U_1,U_2,\ldots,U_\texttt{U}\}$ within a specific time interval. In addition, $\texttt{G}$ \glspl{gcs} $\mathcal{G}\doteq\{G_1,G_2,\ldots,G_\texttt{G}\}$ control the swarm of \glspl{uav}. A mission plan consists of a tuple composed of the assignments of tasks to \glspl{uav}, the orders in which these tasks are performed (that can be represented as a permutation of orders), the assignments of \glspl{uav} to the \glspl{gcs} controlling them, the assignment of sensors that the \glspl{uav} use at each task they perform, and the assignments of the flight profiles used by the \glspl{uav} in their path to each task they perform and in their return to the base.

\begin{figure}[!h]
\begin{center}
\includegraphics[width=\textwidth]{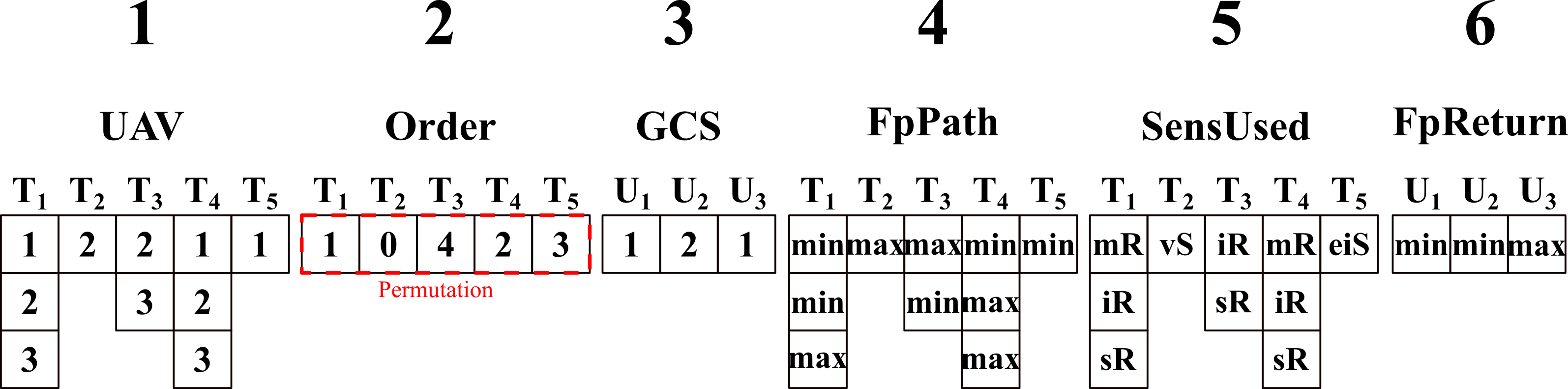}
\caption{Representation of the assignments that compose a mission plan.}
\label{fig:missionplan}
\end{center}
\end{figure}

An example of these plans can be seen in Figure \ref{fig:missionplan}, where there are 5 tasks, 3 \glspl{uav} and 2 \glspl{gcs}. Some tasks are assigned to several \glspl{uav}, which divide the task performance between them, like occurs in tasks 1, 3 and 4. In this example, two types of flight profile are considered: a minimum consumption profile (min) and a maximum speed profile (max). On the other hand, different sensors can be used to perform the tasks, including maritime radars (mR), synthetic aperture radars (sR), inverse synthetic aperture radars (iR) or electro-optical or infrarred sensors (eiS).

When solving a \gls{mcmpp}, there exists several constraints  that must be checked at each solution in order to assure if they are valid, including fuel constraints, time dependencies, path constraints, sensor constraints, \gls{gcs} constraints, distance to ground and between \glspl{uav}, etc. All these constraints have been detailed in previous works~\citep{Ramirez-Atencia2018Constrained,Ramirez-Atencia2018Weighted}.

All of these constraints are defined using a \gls{csp} representation, which is then used in the corresponding algorithm to check the consistency and correctness of the solutions generated. In previous approaches~\citep{Ramirez-Atencia2017Knee}, a \gls{moea} algorithm was used to look for the solutions of this problem.

In this approach, after the solutions have been found, they are treated by the \gls{dss}. In this context, the decision making process consists of selecting the best plan for the mission among the plans returned by the mission planner. To do that, the \gls{dss} consist of two modules: ranking and filtering.

The ranking module sorts the plans returned by the algorithm so the first one should be the best according to the criteria considered, but the rest are also returned to the user so he/she can take the final decision. The filtering module, on the other hand, reduces the number of solutions returned by omitting similar solutions.

To present these solutions to the operator, in order to allow the selection of the best one, a \gls{hmi} was developed in a previous work~\citep{Ramirez-Atencia2018Extending}, so the ordered and filtered solutions are returned in a table indicating the values obtained for each criteria, and then a graphical representation of each of the plans can be visualized in a GIS interface.

\subsection{Mission Planning solutions through Multi-Objective Optimization}\label{mompp}
In previous works~\citep{Ramirez-Atencia2017Knee}, the \gls{mcmpp} was solved using a \gls{moea}; concretely, an extension of \gls{nsga2}~\citep{Deb2002Fast} that use the \gls{csp} model inside the fitness function in order to check if the solutions are valid.

In addition, when solutions are valid, this algorithm considers the multiple optimization objectives of the problem in order to select the fittest solutions that are inside the \gls{pof}. The optimization objectives considered in this problem are the total cost of the mission, the end time of the mission or makespan, the total fuel consumption, the total flight time, the total distance traversed, the risk of the mission and the number of \glspl{uav} used in the mission.

With the high number of optimization objectives considered, the number of optimal solutions of the \gls{pof} becomes huge. As at some point the operator will have to select a concrete plan to be executed, getting so many solutions will highly increase the decision making process. Moreover, some of these non-dominated solutions may not be optimal enough, e.g. a little improvement in some objectives may suppose a big lost in the rest of objectives.

To get rid of this problem, the approach focuses the search on knee points, so only a limited set of the most significant solutions are returned, and not the entire \gls{pof} as other \glspl{moea} usually do. This approach changes the concept of domination, used by \gls{nsga2} to guide the search towards the \gls{pof}, to conedomination, which focus the search on knee points using a specific cone angle that increase the number of solutions that are dominated by a concrete one. With this the number of solutions is highly reduced while the decrease of hypervolume (measuring the quality of the \gls{pof}) is not very high.

With this, the \gls{dss} will receive only the most significant solutions, so the decision making process will be easier for the operator.

\subsection{Ranking system}\label{ranking}
Once the set of plans is obtained, it is necessary to rank them before they are presented to the operator in order to ease the selection process, assuring that most times the final solution selected by the operator is the first in this ranking. In order to do that, several criteria are considered:

\begin{enumerate}
\item The \textbf{makespan}, i.e. the end time of the mission. We have computed it as the maximum of the return times of the \glspl{uav}, which must be minimized.
\item The \textbf{total cost} of the mission, an objective that must always be minimized. To compute it, it is necessary to sum the cost of each \gls{uav} usage, which is computed as the flight time of each \gls{uav} multiplied by the cost per hour of the \gls{uav}.
\item The \textbf{total fuel consumed}, i.e the sum of the fuel consumed by each \gls{uav} at performing the tasks of the mission. This is the main objective in most \gls{uav} mission planning problems that we need to minimize.
\item The \textbf{total distance traversed}, i.e the sum of the distances traversed by each \gls{uav} at performing the tasks of the mission. In most \gls{uav} mission planning problems, this objective is minimized.
\item The \textbf{total flight time}, i.e. the sum of the flight time of each \gls{uav} at performing the tasks of the mission. We have computed it as the sum of the flight times of all the \glspl{uav}, which must be minimized.
\item The \textbf{risk factor for the maximum Percentage of Fuel Usage per \gls{uav}}. It is computed as the maximum of the fuel consumed divided by the initial fuel for all \glspl{uav}, which must be minimized.
\item The \textbf{risk factor for the minimum distance to the ground}, i.e. the minimum of the minimum distances to the ground of each \gls{uav} at performing the tasks of the mission. A percentage indicating the risk is computed with a given risk value, which indicates the maximum risky altitude to the ground. This objective is usually minimized.
\item The \textbf{risk factor for the minimum between \glspl{uav}}, i.e. the minimum of the minimum distances between each pair of \glspl{uav} at performing the mission. A percentage indicating the risk is computed with a given risk value, which indicates the maximum risky distance between \glspl{uav}. This objective is usually minimized.
\item The \textbf{risk factor for the time when \gls{uav} leave \glspl{gcs} coverage zones}. It is computed as the maximum sum of durations in which a \gls{uav} is not covered by the \gls{gcs}, and using a given risk value, a percentage is obtained as in the previous risk factors. This variable must be minimized.
\item The \textbf{number of \glspl{uav} used} in the mission. A mission performed with a lower number of vehicles is usually better because the remaining vehicles can perform other missions at the same time. Therefore, this objective must be minimized.
\item The \textbf{number of \glspl{gcs} used} in the mission. When we have a Multi-\gls{gcs} mission, using a lower number of \glspl{gcs} is usually better in order to use as less resources as possible. Therefore, this objective must be minimized.
\end{enumerate}

In this work, the preferences of the operator define the weights of the criteria needed in the ranking algorithm. This leads to the definition of operator profiles, where each criterion is ranked in degrees of importance, expressed using the following linguistic terms: \emph{Very low (1)}, \emph{Low (2)}, \emph{Medium (3)}, \emph{High (4)} and \emph{Very high (5)}.

These linguistic terms are often fuzzy, and thus, fuzzy weights are used. In this work, \glspl{tfn}\citep{Borovivcka2014} are employed, which are useful to handle imprecise numerical quantities. The membership function used to convert each linguistic term to a \gls{tfn} is presented in Table \ref{tab:fuzzy_weights}. Since in this membership function the distance between numbers is too big and the logical values obtained from the table are critical for the decision maker, it can be said that it is a beginning for all calculations.

\begin{table}[ht]
\centering
\caption{Transformation of linguistic scales for fuzzy membership functions.}
\label{tab:fuzzy_weights}
\begin{tabular}{lc}
  \toprule
  Linguistic scale (Rank) & TFN Membership function \\ 
  \midrule
  Very low & $(0.00, 0.10, 0.25)$ \\ 
  Low & $(0.15, 0.30, 0.45)$ \\
  Medium & $(0.35, 0.5, 0.65)$ \\
  High & $(0.55, 0.70, 0.85)$ \\
  Very high & $(0.75, 0.90, 1.00)$ \\
   \bottomrule
 \end{tabular}
\end{table}

With this, a fuzzy \gls{mcdm} method can be used taking these \glspl{tfn} as weights. This algorithm will be applied to the solutions obtained by the \gls{moea}, which have crisp values. The resultant ordered solutions are then passed to the filtering system.

\subsection{Filtering system}\label{filtering}
Even though a ranked set of solutions is presented, often the operator will check these solutions. In this case, he/she may be overwhelmed if there is a high number of solutions. The filtering of the solutions is performed in order to remove very similar solutions from the final list presented to the operator. Although the knee-point based \gls{moea} presented in section \ref{mompp} already returns a reduced list of solutions, it only looks for similar solutions in terms of optimality, while this filtering system focuses on the similarity of plans, i.e. how similar the assignments are.

\begin{figure}[!h]
	\includegraphics[width=\textwidth]{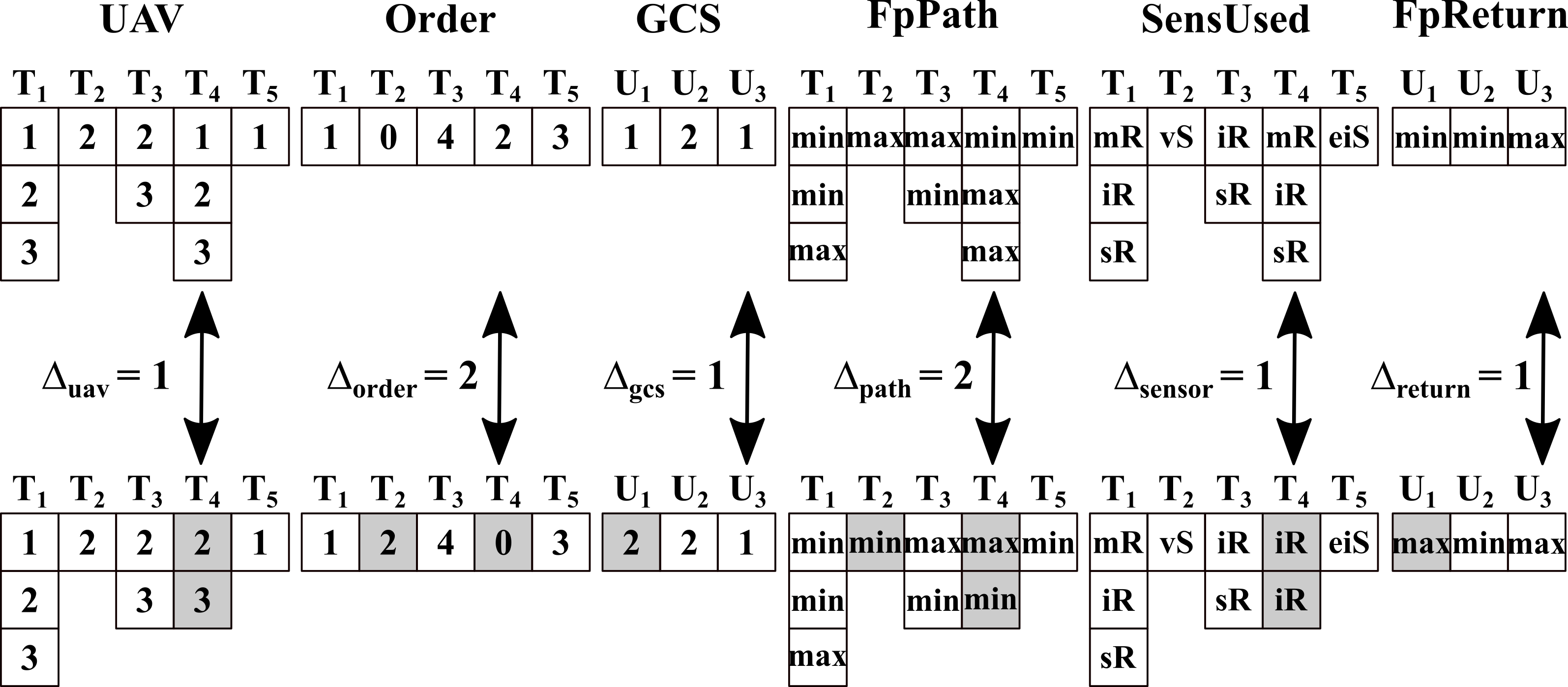}
	\centering
	\caption{Representation of two similar solutions and the filtering values used when applying the distance function}
	\label{fig:diff}
\end{figure}

In order to know the similarity of the solutions, a distance function has been designed. This distance function computes the proximity between two solutions by counting the different values for each variable of the solution, as shown in Figure \ref{fig:diff}, where the different values in the solution below are represented in gray. In this example, $\Delta_{uav}(s_1,s_2)=1$, as the only difference is that \gls{uav} 1 is not performing task $T_4$. Nevertheless, $\Delta_{path}(s_1,s_2)=2$, as flight profiles in the paths for tasks $T_2$ and $T_4$ changed.

The filter distance function needs to define a set of filtering weights for each variable, which should be given according to the degree of importance of that variable. According to Airbus Defence and Space expert operators, the less important variables are those related with flight profiles ($w_{path}$ and $w_{return}$) and sensors used ($w_{sensor}$), so they should be given a low value (e.g. $0.1$), while task assignments ($w_{uav}$) are the most important variables when differentiating two solutions, so they should be given a high value (e.g. $1.0$). To complete the example, Order variables can be given a filtering value of $w_{order}=0.6$ and GCS variables, not so important, a filtering value of $w_{gcs}=0.2$.

So, once we got for each variable the $\Delta$ values, as shown in Figure \ref{fig:diff}, we can compute the planning distance $d_{plan}$ between two solutions as:

\begin{multline}
d_{plan}(s_1,s_2)= w_{uav} \cdot \Delta_{uav}(s_1,s_2) + w_{order} \cdot \Delta_{order}(s_1,s_2) + w_{gcs} \cdot \Delta_{gcs}(s_1,s_2) +\\
 w_{path} \cdot \Delta_{path}(s_1,s_2) + w_{return} \cdot \Delta_{return}(s_1,s_2) + w_{sensor} \cdot \Delta_{sensor}(s_1,s_2)
\end{multline}

Then, a threshold value is used to decide when two solutions are too similar and one of them must be removed from the set of solutions. The removed solution will be the one with the lowest ranking value. The selection of the threshold value must provide the biggest deletion of similar solutions but maintaining the quality of the \gls{pof}.

\section{Experiments}\label{experiments}
In order to evaluate the proposed \gls{dss}, both its ranking and filtering systems, 12 missions of increasing complexity were applied the knee-point based \gls{moea} developed in a previous work~\citep{Ramirez-Atencia2017Knee}, and different solutions were obtained. Table \ref{tab:datasets} presents the complexity factors of these missions and the number of solutions obtained.

\begin{table}[!h]
\caption{Features of the different datasets designed.}
\label{tab:datasets}
\begin{center}
\begin{tabular}{|c!{\vrule width 3pt}c|c|c|c!{\vrule width 3pt}c| }
  \hline
  \begin{minipage}[t]{1cm}
  \centering
  Dataset \\
  Id.
  \end{minipage} & Tasks & \begin{minipage}[t]{1.8cm}
  \centering
  Multi-UAV \\
  Tasks
  \end{minipage} & UAVs & GCSs & \begin{minipage}[t]{2cm}
  \centering
  Number of \\
  Solutions
  \end{minipage}\\[3ex]
  \noalign{\hrule height 2pt}	
  1 & 6 & 1 & 3 & 1 & 17 \\
  \hline
  2 & 6 & 1 & 4 & 2 & 5 \\
  \hline
  3 & 8 & 2 & 5 & 2 & 38 \\
  \hline
  4 & 9 & 2 & 5 & 2 & 9 \\
  \hline
  5 & 9 & 2 & 6 & 2 & 5 \\
  \hline
  6 & 10 & 2 & 6 & 2 & 18 \\
  \hline
  7 & 11 & 3 & 6 & 2 & 3 \\
  \hline
  8 & 12 & 3 & 7 & 3 & 11 \\
  \hline
  9 & 12 & 3 & 8 & 3 & 5 \\
  \hline
  10 & 13 & 4 & 7 & 3 & 4 \\
  \hline
  11 & 14 & 4 & 8 & 3 & 5 \\
  \hline
  12 & 16 & 5 & 10 & 3 & 8  \\
  \hline
\end{tabular}
\end{center}
\end{table}

Then, in order to select the best \gls{mcdm} method for ranking, 6 fuzzy \gls{mcdm} methods were considered: Fuzzy \gls{ahp}, Fuzzy VIKOR, Fuzzy \gls{topsis} (where two versions based on different normalisation procedures are considered: Fuzzy TOPSISVector, using the vector normalisation, and Fuzzy TOPSISLinear, using the linear transformation of maximum), Fuzzy \gls{mmoora} and Fuzzy \gls{waspas}. In addition, 10 classical \gls{mcdm} methods were also considered in the comparison: \gls{wsm}, \gls{wpm}, \gls{ahp}, VIKOR, \gls{topsis} (both TOPSISVector and TOPSISLinear), ELECTRE III, \gls{mmoora}, \gls{rim} and \gls{waspas}.

In these methods, the linguistic scale presented in section \ref{ranking} is used with the fuzzy set transformation of Table \ref{tab:fuzzy_weights} for the fuzzy \gls{mcdm} methods. But for classical \gls{mcdm} methods, these linguistic terms are converted into numeric weights. So the weights can be expressed as a percentage factor, so for each criteria factor $f \in F$, given its degree of importance $D_I(f) \in \{1,2,3,4,5\}$, its corresponding weight is computed as:

\begin{equation}
    w(f) = \frac{D_I(f)}{\sum_{f' \in F}{D_I(f')}}
\end{equation}

In the case of \gls{ahp}, the pairwise comparison matrix of criteria is constructed based on the different scales for each criterion on the operator profile. Each pairwise value for $f, f' \in F$ is $D_I(f)/D_I(f')$. Similarly, for the pairwise comparison matrix of alternatives w.r.t a given criterion, the performance values $v_f$ for each pair of alternatives $s, s' \in S$ are converted into a 1-9 scale based on the minimum and maximum values of that criteria (obtained in the \gls{moea}). In the fuzzy case, both weights and alternatives pairwise comparisons are transformed into \glspl{tfn}. This does not apply when two alternatives have the same value, where the crisp number will be used instead.

For ELECTRE III, the indifferent, preference and veto threshold were defined by expert operators. These values are presented in Table \ref{tab:thresholds}. In \gls{rim}, the range $[A,B]$ for each criterion is set to the minimum and maximum values obtained in the \gls{moea}, while the reference ideal $[C,D]$ is set to the best values obtained for each criterion in an interval with the indifference threshold. For the rest of parameters, we define $v=0.5$ in VIKOR and $\lambda=0.5$ in \gls{waspas}.

\begin{table}[ht]
\centering
\caption{Indifferent ($q$), Preference ($p$) and Veto ($v$) Thresholds for ELECTRE III used in this experiment.}
\label{tab:thresholds}
\begin{tabular}{l|ccc}
  \cline{2-4}
  & $q$ & $p$ & $v$ \\ 
  \midrule
  Cost & 0.5 & 5.0 & 50.0 \\ 
  Distance (km) & 0.5 & 5.0 & 50.0 \\ 
  Flight Time (h) & 0.01 & 0.5 & 1.5 \\ 
  Fuel (kg) & 0.5 & 7.0 & 50.0 \\ 
  Makespan (h) & 0.005 & 0.3 & 1.0 \\ 
  Num GCSs & 0 & 1 & 2 \\ 
  Num UAVs & 0 & 1 & 4 \\ 
  Risk Distance Ground (\%) & 0.001 & 0.1 & 0.5 \\ 
  Risk Distance UAVs (\%) & 0.001 & 0.1 & 0.5 \\ 
  Risk Fuel Usage (\%) & 0.001 & 0.1 & 0.5 \\ 
  Risk Out of Coverage (\%) & 0.001 & 0.1 & 0.5 \\ 
  \bottomrule
 \end{tabular}
\end{table}

In addition, six operator profiles were created considering different importance degrees on each optimization criterion (very low, low, medium, high, very high). These profiles are shown in Table \ref{tab:profiles}. Then, these values are used as weights through the aforementioned transformations. 

\begin{table}[ht]
\centering
\caption{Operator Profiles used in this experiment.}
\label{tab:profiles}
\scalebox{0.8}{\begin{tabular}{l|llllll}
  \cline{2-7}
  & Balanced & Cost & Time & Risk & Resources & RiskCost \\ 
  \midrule
Cost & Medium & Very High & Medium & Low & High & Very High \\ 
  Distance & Medium & Medium & Medium & Low & Medium & Medium \\ 
  Flight Time & Medium & Low & High & Medium & Low & Low \\ 
  Fuel & Medium & High & Medium & Medium & High & High \\ 
  Makespan & Medium & Low & Very High & Medium & Low & Low \\ 
  Num GCSs & Medium & Medium & Medium & High & Very High & Medium \\ 
  Num UAVs & Medium & High & Medium & High & Very High & High \\ 
  Risk Distance Ground & Medium & Medium & Low & Very High & Medium & Very High \\ 
  Risk Distance UAVs & Medium & Medium & Low & Very High & Medium & Very High \\ 
  Risk Fuel Usage & Medium & Medium & Low & Very High & Medium & Very High \\ 
  Risk Out of Coverage & Medium & Low & Low & Very High & Medium & Very High \\ 
   \bottomrule
 \end{tabular}}
\end{table}

\subsection{Evaluation Criteria}
In order to perform an external evaluation of the quality of a ranking, and to compare different ranking systems objectively, we have created a ``ground truth" dataset based on collective human judgement. Human judgement as a way to create ground truth data is a common approach used in many domains such as image recognition or performance analysis \citep{Rodriguez-Fernandez2017Analysing}.

In this work, the evaluation (or ground truth) data is created by asking 4 expert \gls{uav} operators to decide, given a set of unranked mission plans of the same scenario and a concrete operator profile, which of them is the best, i.e., which of them they would choose to be executed in a real environment. For each decision submitted by the operator, the following data is saved:
\begin{itemize}
\item The operator profile.
\item An identifier of the mission scenario being evaluated.
\item An identifier of the chosen plan.
\end{itemize}

The process of gathering human judgement-based evaluation data has been automated by the use of a web-app, whose graphical user interface can be seen in Figure \ref{fig:dssevaluationapp}. The use of this app can be summarised in three iterative steps:

\begin{figure}[!h]
\fbox{\includegraphics[width=\textwidth]{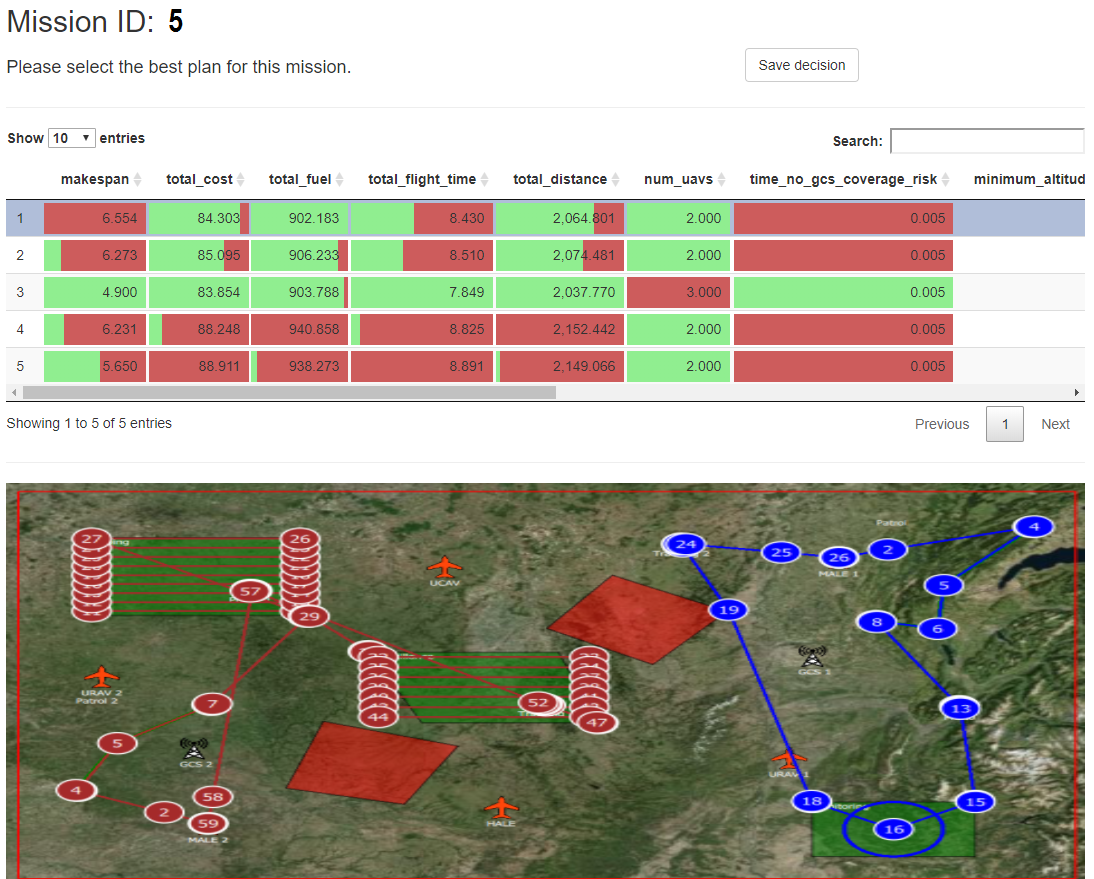}}
\caption{Screenshot of the app developed to retrieve the evaluation data.}
\label{fig:dssevaluationapp}
\end{figure}

\begin{enumerate}
\item The operator selects an operator profile. These profiles taking part of the evaluation process must be loaded beforehand.

\item For each mission scenario to evaluate, a table with several possible solutions (plans) is displayed, one per row. Each column shows the value of a specific risk factor. To allow an easier comparison between the solutions, each cell is shaded with a two-coloured progress bar. The green part of the bar represents the relative quality of the risk factor with respect to the rest of solutions in the table. The red part is only used to fill the rest of the cell. As an example, the solution(s) with the minimum makespan in the table will display the corresponding makespan cell with a fully green shade. On the contrary, the solution(s) that maximizes the makespan will be fully red. In addition to the table, the operator may also select any of the solutions to visually inspect the plan in the mission map (See Figure \ref{fig:dssevaluationapp}).

\item The operator judges which is the best plan by selecting it on the table and clicking the ``Save decision" button. The app will automatically restart the decision process with a new mission.
\end{enumerate}

With the evaluation dataset created, we can give a score to the rankings created by a \gls{mcdm} algorithm in the context of this work. The score is aimed to assess whether the best plan selected by the operator is found within the top results of the algorithm or not. Let $a$ be a \gls{mcdm} algorithm, $m$ a mission scenario and $\textnormal{op}$ an operator profile, then $L(a,m, \textnormal{op})$ denotes the ordered list of solutions (plans) returned by the algorithm in that context. Having an ordered list $L$, $r(L,p)$ denotes the rank position of plan $p$ in the list $L$, where 1 means ``best". On the other hand, let $\textnormal{op}$ be an operator profile, we refer to $S_{dm}(\textnormal{op},m)$ as the selection of best plan made by the operator or decision maker $dm$ for the profile $\textnormal{op}$ in the context of mission $m$. This selection is retrieved from the evaluation dataset. With all this, we can define the \emph{score} of an algorithm for a given operator profile and a given mission scenario as follows:
\begin{equation}
\label{eq:score}
\textit{Score}_{dm}(a,\textnormal{op},m) = \frac{\textnormal{num\_solutions}(m) - r(L(a,m, \textnormal{op}), S_{dm}(\textnormal{op}, m))}{\textnormal{num\_solutions}(m) - 1},
\end{equation}

As it can be seen, the value of the score is bounded on $[0, 1]$, where $1$ represents the best possible matching (the selected plan is the first of the algorithm ranking) and $0$ the worst (the selected plan is the last of the algorithm ranking). Then, the final score is computed as the average of scores obtained for each operator. In order to give a general score for an algorithm, we take the average value of Eq. \ref{eq:score} over every operator, profile and mission scenario available in the evaluation dataset.

\subsection{Comparing MCDM methods for the ranking system}
Now, the different \gls{mcdm} and fuzzy \gls{mcdm} methods for the ranking system are compared in order to find the best one for the mission plan selection. The experiments have been performed in three steps:

\begin{enumerate}
\item Each algorithm has been executed for every mission with the different operator profiles as weights of the optimization criteria.
\item The operators have selected the best plan using an evaluation app.
\item The score metric is applied to every tuple $\left\langle \textnormal{operator}, \textnormal{mission}, \textnormal{profile}, \textnormal{algorithm} \right\rangle$ obtained after finishing the previous steps.
\end{enumerate}

In order to give a global score for an algorithm, we take the average value of Equation \ref{eq:score} over every mission scenario available in the evaluation dataset and every operator profile. With these results, in order to compare if fuzzy \gls{mcdm} methods are better than classical \gls{mcdm} methods, Table \ref{tab:comparison_mcdm} shows the difference between the average scores obtained for each pair of fuzzy and non-fuzzy \gls{mcdm} method.

\begin{table}[ht]
\centering
\caption{Comparison between classical and Fuzzy MCDM methods in the context of ranking solutions of a multi-UAV mission planning problem. Cells represent the difference in the average score between fuzzy methods (columns) and classical methods (rows). Underlined cells indicate that the difference is statistically significant (p-value $<$ 0.05).} 
\label{tab:comparison_mcdm}
\scalebox{0.8}{
\begin{tabular}{rllllll}
  \hline
 &   \begin{minipage}[t]{0.7cm}
  Fuzzy \\
  AHP
  \end{minipage}  & \begin{minipage}[t]{1cm}
  Fuzzy \\
  VIKOR
  \end{minipage} & \begin{minipage}[t]{2cm}
  Fuzzy \\
  TOPSISLinear
  \end{minipage} & \begin{minipage}[t]{2cm}
  Fuzzy \\
  TOPSISVector
  \end{minipage} & \begin{minipage}[t]{2cm}
  Fuzzy \\
  MultiMOORA
  \end{minipage} & \begin{minipage}[t]{1.4cm}
  Fuzzy \\
  WASPAS
  \end{minipage} \\ 
  \hline
  WSM & \underline{0.042} & \underline{0.166} & \underline{0.097} & 0.033 & \underline{0.073} & \underline{0.092} \\ 
  WPM & -0.03 & \underline{0.094} & \underline{0.025} & \underline{-0.039} & 0.001 & 0.02 \\ 
  AHP & -0.021 & \underline{0.103} & \underline{0.035} & \underline{-0.029} & 0.011 & \underline{0.029} \\ 
  VIKOR & \underline{-0.116} & 0.008 & \underline{-0.06} & \underline{-0.125} & \underline{-0.085} & \underline{-0.066} \\
  TOPSISLinear & -0.008 & \underline{0.116} & \underline{0.047} & \underline{-0.017} & 0.023 & \underline{0.042} \\ 
  TOPSISVector & 0.032 & \underline{0.156} & \underline{0.087} & \underline{0.023} & \underline{0.063} & \underline{0.082} \\ 
  ELECTRE & \underline{0.042} & \underline{0.167} & \underline{0.098} & \underline{0.034} & \underline{0.074} & \underline{0.093} \\
  MultiMOORA & -0.036 & \underline{0.089} & \underline{0.02} & \underline{-0.044} & -0.004 & 0.014 \\ 
  RIM & \underline{0.039} & \underline{0.163} & \underline{0.094} & 0.03 & \underline{0.07} & \underline{0.089} \\ 
  WASPAS & \underline{-0.045} & \underline{0.079} & 0.011 & \underline{-0.054} & -0.014 & \underline{0.005} \\ 
   \hline
\end{tabular}}
\end{table}

In this table, each column represents the difference of score between the fuzzy \gls{mcdm} method of that column and the \gls{mcdm} method of each row. It is clearly appreciable that there are more positive differences, which means that fuzzy methods obtained in general better results than non-fuzzy methods. On the other hand, the Wilcoxon signed-rank test was used to check the statistical significance of these results, where the underlined values obtained a p-value less than $0.05$ for that pair of algorithms. In general, it can be appreciate that fuzzy \gls{mcdm} methods are statistically significant with respect to \gls{mcdm} methods. So, it can be inferred that fuzzy \gls{mcdm} methods are more appropriate to deal with this kind of problem.

\begin{figure}[!h]
\centering
\includegraphics[width=\textwidth]{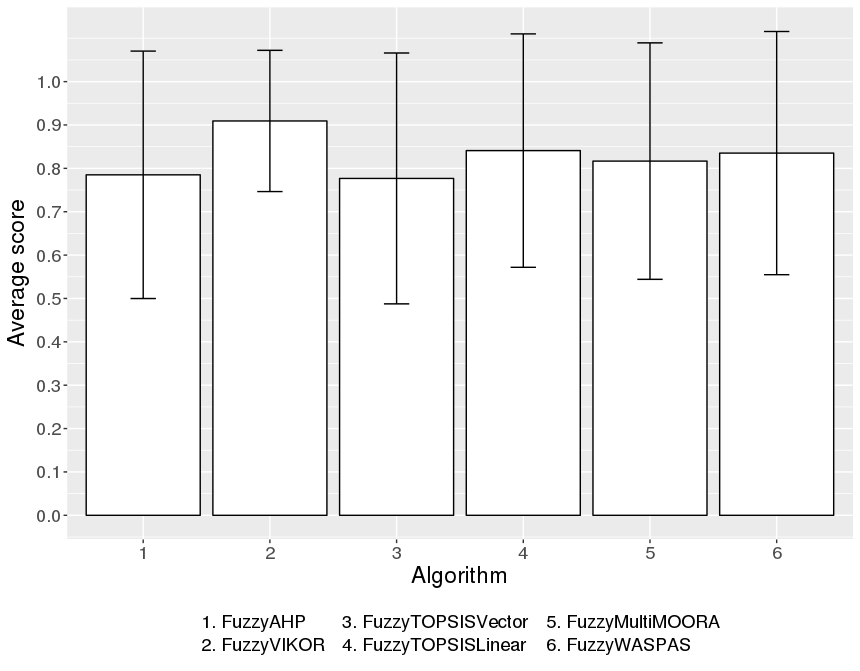}
\caption{Average scores and standard deviation of the different Fuzzy MCDM algorithms applied to the plan selection for MPP.}
\label{fig:fuzzy_comparative}
\end{figure}

Now, focusing on the different Fuzzy \gls{mcdm} methods, Figure \ref{fig:fuzzy_comparative} clearly shows that Fuzzy VIKOR obtained the best results. This approach was validated as the most suitable to apply in the context of this work. In order to check the statistical significance of these results, the Wilcoxon signed-rank test was used again, comparing Fuzzy VIKOR with the rest of fuzzy methods. The p-value obtained was less than $0.05$ for all of the algorithms. It is interesting that Fuzzy \gls{ahp}, although being a quite popular method, did not obtain so good results as the rest. This is probably because the pairwise comparisons were constructed from weights of criteria, and not directly provided by decision makers. As the rest of methods are based on weights, it is not surprising that they obtain better results for this kind of problem.

\begin{figure}[!h]
\centering
\includegraphics[width=0.8\textwidth]{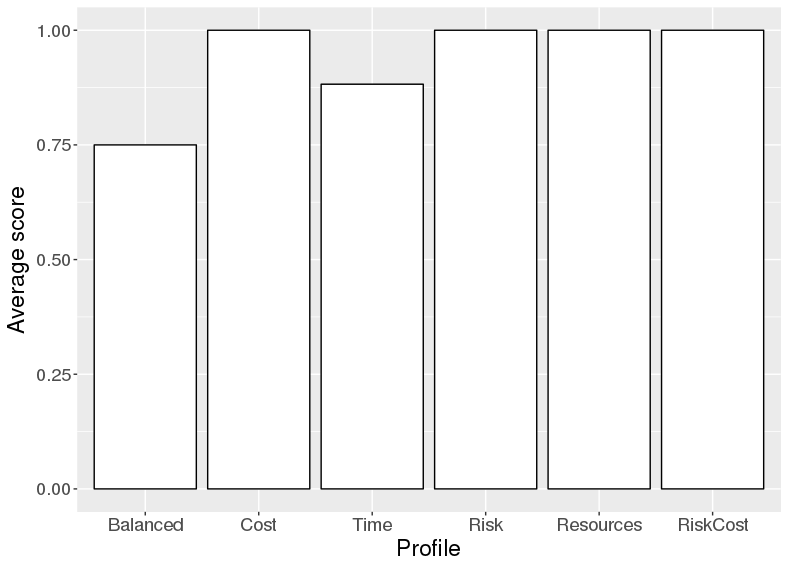}
\caption{Median scores for the different operator profiles used in the plan selection of the MPP.}
\label{fig:algorithm_profile_comparative}
\end{figure}

Now, in order to see if some operator profile was harder for the \gls{mcdm} methods, the median of the scores aggregated by operator profile are presented in Figure \ref{fig:algorithm_profile_comparative}. It is clearly appreciable that \emph{Balanced} is the most difficult profile to evaluate, followed by the \emph{Time} profile. This could be due to the different conception of what balance means to an expert operator and to an algorithm based on fuzzy weights.

In conclusion, the ranking of solutions of the \gls{mcmpp} can be done using Fuzzy VIKOR, which obtained the best performance with the smallest variance in the experimental results. An optimal result is obtained with most operator profiles, except for balanced profiles, which are harder to solve for most \gls{mcdm} algorithms.

\subsection{Tuning the threshold for the filtering system}
In order to decide which is the best possible threshold value for the \gls{mcmpp}, the ranked plans obtained in the previous section were used. With this, we apply the similarity function $d_{plan}(s_1,s_2)$ to each pair of solutions $s1$ and $s2$.

\begin{figure}[t]
	\centering
	\includegraphics[width=\textwidth]{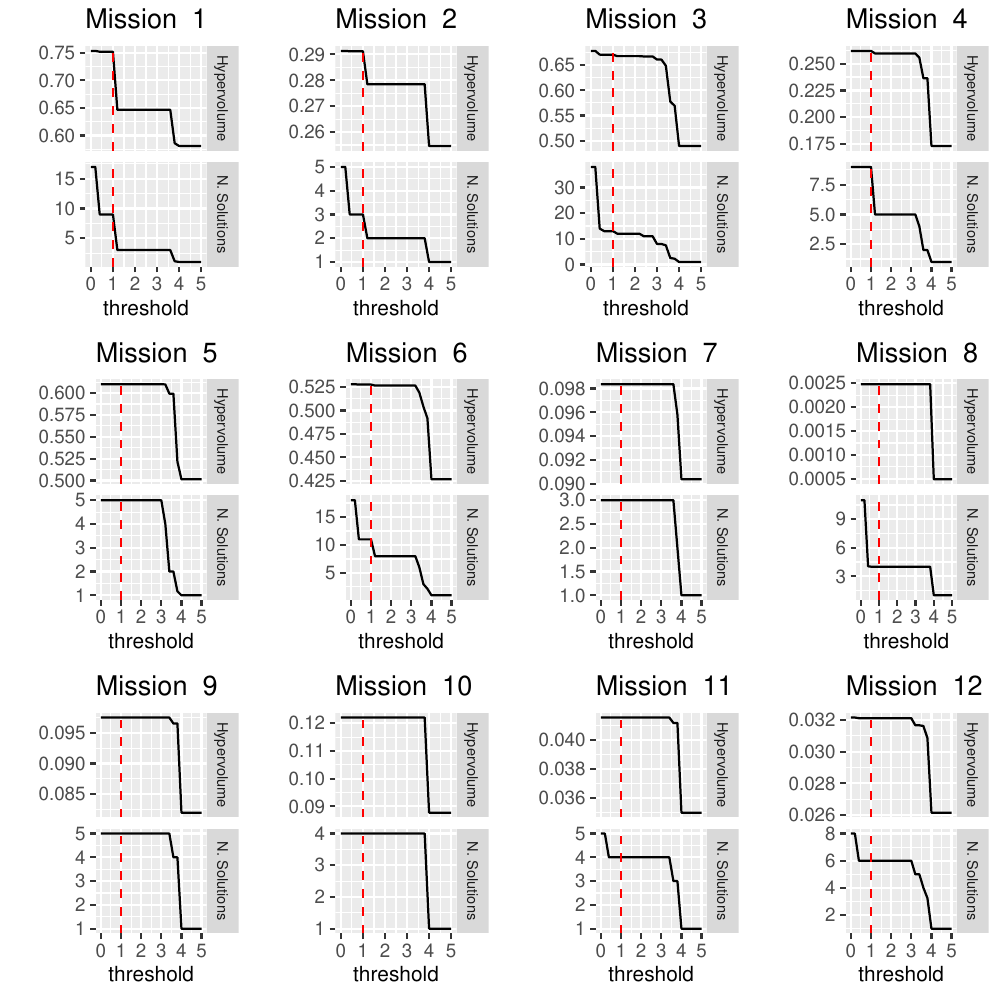}
	\caption{Hypervolume and number of remaining solutions after the similarity filtering with different threshold values. The red line marks the common threshold chosen as most suited for this experimentation after an empirical study.}
	\label{fig:filter}
\end{figure}

Then, different threshold values from 0 to 5, with an increment of 0.1, are tested, and the remaining solutions after the filtering using that threshold are used to compute the hypervolume of these filtered sets. The results obtained are shown in Figure \ref{fig:filter}, where each plot represents a mission of the problem, the threshold is presented in the horizontal axis and the hypervolume and the number of remaining solutions are presented in the vertical axis. With this, it can be concluded that the best threshold value (for \glspl{mcmpp}) to be used is 1 (see the red line in the figure), since missions 1 and 2 have a great loss of hypervolume for bigger values than 1. Looking at the rest of missions, this value could be bigger, e.g. 3, since the hypervolume does not highly decrease before this value, but neither does the number of solutions, which are pretty similar between threshold values of 1 and 3.

\section{Conclusions and Future Work}\label{conclusions}
In this work, a \gls{dss} for plan selection in the context of the \gls{mcmpp} has been designed and tested. This \gls{dss} requires the solutions obtained by a mission planner, and the operator profiles, which can be directly specified by the operators, where each criterion of the decision problem is given a degree of importance using a linguistic term.

The \gls{dss} presented is composed of two modules: the ranking module and the filtering module. In the first module, the solutions are ranked based on the operator profiles using a fuzzy \gls{mcdm} algorithm. Several fuzzy and non-fuzzy \gls{mcdm} algorithms were tested with the solutions obtained in a previous work, and it was concluded that fuzzy methods got better results than non-fuzzy methods; and, for this problem, Fuzzy VIKOR was the fittest one.

On the other hand, the filtering module uses the ranked solutions returned by the ranking module and filters those which are more similar based on a distance function. This function compares each variable of the \gls{csp} model of the \gls{mcmpp}, providing a specific weight depending on the importance of the variable. In order to decide the threshold value of the filtering function, different thresholds were tested, and it was concluded that the value $1$ was the fittest value in terms of highest hypervolume and lowest number of remaining solutions.

This \gls{dss} highly decreases the operator workload since it facilitates the operator decision process, which is pretty hard when the number of solutions returned is large. In any case, although the operator is provided with the final list of ranked and filtered solutions, the final selection that must be eventually made, depends on his/her decision. For this, it is necessary to provide a well-designed environment that allows the operator to visualize the ranked alternatives and select the most appropriate. This is described in the following chapter.

In future works, it would be interesting to improve the system by considering the final selection made by the operators. Using this and some pattern recognition or machine learning techniques, the operator profile would be updated, aiming to perform better in future decisions. On the other hand, some machine learning technique could be used to automatize the threshold selection for the filtering system according to the plans obtained.

\section*{Acknowledgments}
This work has been supported by Airbus Defence \& Space (FUAM-076914 and FUAM-076915), and by the next research projects:  DeepBio (TIN2017-85727-C4-3-P), by Spanish Ministry of Economy and Competitivity, and CYNAMON (CAM grant S2018/TCS-4566), under the European Regional Development Fund FEDER. The authors would like to acknowledge the support obtained from Airbus Defence \& Space, specially from Savier Open Innovation project members: Jos\'e Insenser, Gemma Blasco and C\'esar Castro.

\bibliography{mybibfile}

\end{document}